\documentclass[10pt,twocolumn,letterpaper]{article}

\usepackage{cvpr}              
\usepackage{epsfig}
\usepackage{graphicx}
\usepackage{amsmath}
\usepackage{amssymb}

\usepackage[belowskip=-5pt,aboveskip=3pt]{caption} 

\usepackage{multirow}
\usepackage{tabularx}
\usepackage{color}
\usepackage{amsthm}
\usepackage[export]{adjustbox}
\usepackage{comment}

\usepackage[utf8]{inputenc} 
\usepackage[T1]{fontenc}    
\usepackage{booktabs}       
\usepackage{amsfonts}       
\usepackage{nicefrac}       
\usepackage{microtype}      
\usepackage{enumitem}
\usepackage{blindtext}
\usepackage{sidecap}

\newcommand{\highlight}[1]{\textbf{#1.}}

\definecolor{citecolor}{RGB}{65,105,225}
\usepackage[breaklinks=true,colorlinks,citecolor=citecolor,bookmarks=false]{hyperref}

\usepackage{dsfont}
\definecolor{dg}{rgb}{0,0.694,0.298}
\definecolor{purple}{rgb}{0.4,0.176,0.569}
\definecolor{royalblue}{RGB}{65,105,225}
\usepackage{pifont}
\newcommand{\figref}[1]{Fig.~\ref{#1}}
\newcommand{\reqref}[1]{Eq.~\eqref{#1}}
\newcommand{\secref}[1]{Sec.~\ref{#1}}
\newcommand{\tableref}[1]{Table~\ref{#1}}

\makeatletter
\DeclareRobustCommand\onedot{\futurelet\@let@token\@onedot}
\def\@onedot{\ifx\@let@token.\else.\null\fi\xspace}
\def\eg{\emph{e.g}\onedot} 
\def\ie{\emph{i.e}\onedot}

\def\etal{\emph{et al}\onedot}
\makeatother

\definecolor{americanrose}{rgb}{1.0, 0.01, 0.24}

\newcommand{\topone}[1]{\textbf{\textcolor{red}{#1}}}

\def\cvprPaperID{4406} 
\def\confName{CVPR}
\def\confYear{2022}

\begin{document}

\title{MISF:Multi-level Interactive Siamese Filtering for High-Fidelity Image Inpainting}


\author{Xiaoguang Li\textsuperscript{1}\thanks{Xiaoguang Li and Qing Guo are co-first authors and contribute equally.}, 
~Qing Guo\textsuperscript{2\rm$*$},
~Di Lin\textsuperscript{3},
~Ping Li\textsuperscript{4},\\
Wei Feng\textsuperscript{3},
~Song Wang\textsuperscript{1}\\~\\
\textsuperscript{1}University of South Carolina, USA,~~ 
\textsuperscript{2}Nanyang Technological University, Singapore \\
\textsuperscript{3}Tianjin University, China, ~~
\textsuperscript{4}Hong Kong Polytechnic University, China
}

\twocolumn[{%
\renewcommand\twocolumn[1][]{#1}%
\maketitle
\begin{center}
\centering
\includegraphics[width=1.0\textwidth]{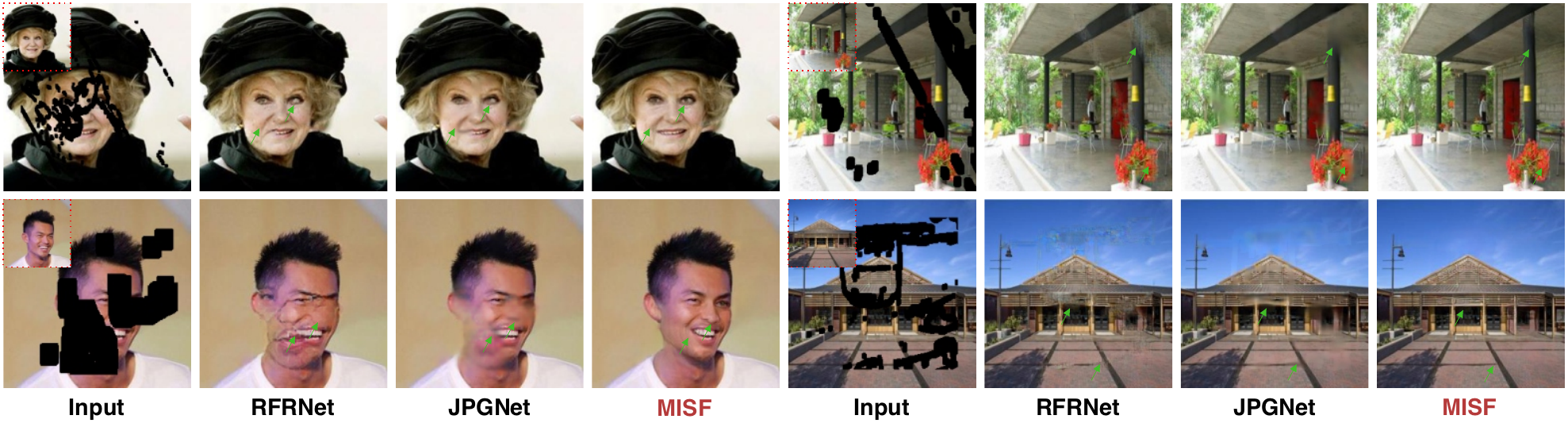}
\captionof{figure}{Four examples of using state-of-the-art methods (\ie, RFRNet \cite{li2020recurrenta} and JPGNet \cite{guo2021jpgnet}) and the proposed method for image inpainting. Our method is able to complete the missing pixels and produce realistic and high-fidelity images. We highlight the main differences via green arrows. We show the ground truth at the upper left corner of the input images.}
\label{fig:mot}
\end{center}
}]


\begin{abstract}
Although achieving significant progress, existing deep generative inpainting methods are far from real-world applications due to the low generalization across different scenes. As a result, the generated images usually contain artifacts or the filled pixels differ greatly from the ground truth.
Image-level predictive filtering is a widely used image restoration technique, predicting suitable kernels adaptively according to different input scenes.
Inspired by this inherent advantage, we explore the possibility of addressing image inpainting as a filtering task.
To this end, we first study the advantages and challenges of the image-level predictive filtering for image inpainting: the method can preserve local structures and avoid artifacts but fails to fill large missing areas.
Then, we propose the \textit{semantic filtering} by conducting filtering on deep feature level, which fills the missing semantic information but fails to recover the details.
To address the issues while adopting the respective advantages, we propose a novel filtering technique, \ie, \textit{Multi-level Interactive Siamese Filtering (MISF)}, which contains two branches: kernel prediction branch (KPB) and semantic~\&~image filtering branch (SIFB).
These two branches are interactively linked: SIFB provides multi-level features for KPB while KPB predicts dynamic kernels for SIFB. As a result, the final method takes the advantage of effective semantic~\&~image-level filling for high-fidelity inpainting.
Moreover, we discuss the relationship between MISF and the naive encoder-decoder-based inpainting, inferring that MISF provides novel \textit{dynamic convolutional operations} to enhance the high generalization capability across scenes.
We validate our method on three challenging datasets, \ie, Dunhuang, Places2, and CelebA. Our method outperforms state-of-the-art baselines on four metrics, \ie, $L_1$, PSNR, SSIM, and LPIPS. Please try the released code and model in \url{https://github.com/tsingqguo/misf}.
\end{abstract}

\section{Introduction}

Image inpainting is a fundamental problem in computer vision and artificial intelligence applications. The main goal is to fill missing pixels in an image and make it identical to the clean one.
Recent works mainly address the task by modeling it as a generation task \cite{yeh2017semantic,pathak2016context, liu2018image, li2020recurrenta}.
As a result, they can employ cutting-edge deep generative techniques (\eg, generative adversarial network \cite{goodfellow2014generative,mirza2014conditional}) to realize high-quality restoration on challenging datasets. However, the generative network-based inpainting encodes the input image to a latent space and then decodes it to a new image. Such a process neglects the explicit prior, \ie, smoothness across neighboring pixels or features, and the fidelity of inpainting fully relies on the data and training strategy.

Note that, different from the generation task, image inpainting has its specific challenges:
\textit{First}, the image inpainting requires the completed images to respect the clean image (\ie, to produce high-fidelity images) and to be natural. These requirements make image inpainting different from the pure image generation task that mainly focuses on naturalness.
\textit{Second}, the missing areas' shapes may be different and the background scenes are diverse. These facts require the inpainting method to have high generalization capability across missing areas and scenes.
Although deep generative networks achieve significant progress on image inpainting, they are far from solving the above challenges.
For example, the recent work RFRNet \cite{li2020recurrenta} conducts feature reasoning on the encoder-decoder network and achieves state-of-the-art performance on public datasets.
Nevertheless, given different faces with different missing areas, it is hard to produce high-fidelity inpainting results. Moreover, the artifacts appear in the results.
As shown in the \figref{fig:mot}, when addressing the first case with small missing areas, RFRNet can generate a natural face. However, when comparing with the ground truth, we see that the local structures around the arrows are distorted.
In terms of the second case with larger missing areas, RFRNet even fails to produce a natural face.
When handling other natural scenes (\eg, the third and fourth cases), RFRNet also introduces small artifacts.

Guo \etal \cite{guo2021jpgnet} have noticed above issues of generative-based inpainting methods \cite{nazeri2019edgeconnect,ren2019structureflow,li2020recurrenta,guo2021jpgnet} and propose JPGNet that uses image-level predictive filtering to alleviate the artifacts.
The image-level predictive filtering reconstructs pixels via their neighboring pixels. The filtering kernels are adaptively estimated according to the inputs.
As a result, JPGNet can recover the local structure while avoiding the artifacts, thus helping RFRNet achieve significant quality improvement.
As shown in \figref{fig:mot}, the artifacts of RFRNet are smoothed effectively.
Nevertheless, many details are lost, while the real structures fail to be recovered.

Inspired by the inherent advantages of predictive filtering on adaptiveness and restoration, we propose a novel framework to handle the two challenges.
Specifically, we have three main efforts: \textit{First}, we study the advantages and challenges of adopting the existing predictive filtering method for image inpainting, that is, the image-level predictive filtering can restore local structures and avoid artifacts but cannot fill large missing areas.
\textit{Second}, we extend the image-level filtering to the deep feature level and propose the \textit{semantic filtering}, which can complete large missing areas but loses details.
\textit{Third}, to address the issues, we propose a novel filtering technique, \ie, \textit{Multi-level Interactive Siamese Filtering (MISF)}, which contains two branches: kernel prediction branch (KPB) and semantic~\&~image filtering branch (SIFB).
These two branches are interactively linked at semantic~\&~pixel levels. SIFB provides multi-level features for KPB while KPB predicts dynamic kernels for SIFB. MISF can utilize the smoothness prior across neighbors explicitly and reconstruct clean pixels or features by linearly combining the neighbors. As a result, the final method takes the advantage of effective semantic~\&~pixel-level filling for high-fidelity inpainting.
As shown in \figref{fig:mot}, our method can generate natural and high-fidelity images under different scenes with different missing areas.
In addition, we conduct an insight discussion about the relationship between our method and the naive generative network, inferring that our method is multi-level \textit{dynamic convolutional operations} that adjusts the convolutional parameters according to different inputs, which lets our method have high generalization.
We conduct extensive experiments on three challenging datasets (\ie, Place2, CelebA, and Dunhuang) and achieve much better scores than the competitive methods on the public datasets in terms of four quality metrics.

\section{Related Work}

%
\begin{figure}
\centering
\includegraphics[width=1.0\linewidth]{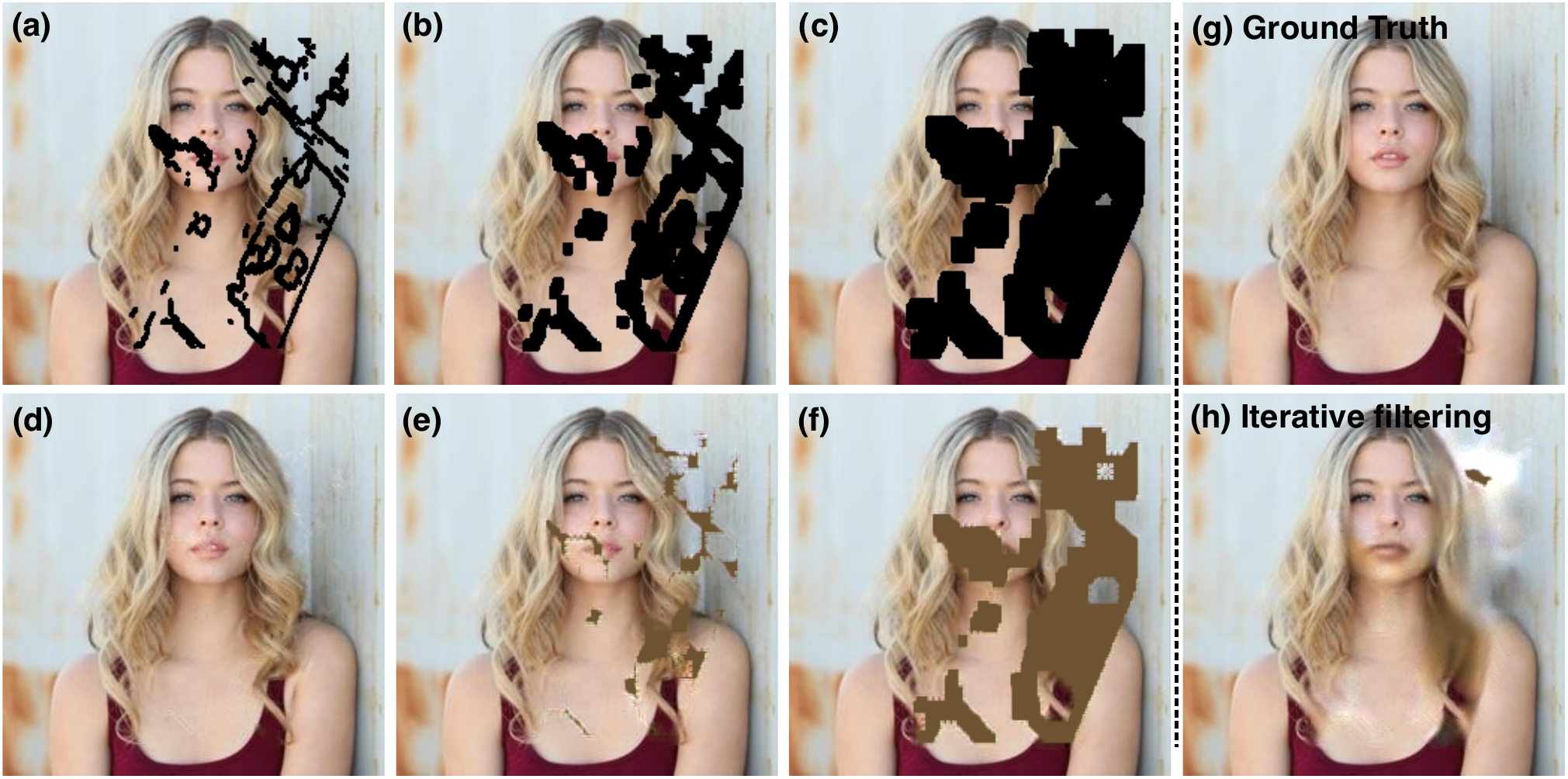}
\caption{
 Using predictive filtering for image inpainting under three mask sizes. The images (a), (b), and (c) are fed to the predictive filtering and we get (d), (e), and (f), respectively. We also try to complete the image (c) via the predictive filtering recurrently and obtain (h).
}
\label{fig:observation}
\vspace{-15pt}
\end{figure}
%

\highlight{Deep generative networks for image inpainting}
The conventional image inpainting methods \cite{barnes2009patchmatch, bertalmio2003simultaneous, chan2001nontexture, ding2018image, fang2019efficient, li2016image, song2019geometry} focus on finding useful patches for recovering the damaged image regions. However, the semantic information of image regions is out of consideration in these methods, thus yielding unsatisfactory results in complex scenes.


More recent methods employ deep generative adversarial networks \cite{goodfellow2014generative} to learn semantic information from data for better inpainting. Pathak et al. \cite{pathak2016context} and Iizuka et al. \cite{iizuka2017globally} use the conditional GAN \cite{mirza2014conditional}, along with the powerful information encoder, for a better polish of the image details in the inpainting result. Moreover, Iizuka et al. \cite{iizuka2017globally} enforce the local and global consistency of the recovered regions. Li et al. \cite{li2020recurrenta} propose the recurrent reconstruction of the corrupted image on the convolutional feature of an image. In addition, Yan et al. \cite{yan2018shift} and Yu et al. \cite{yu2018generative} propose the contextual model for capturing the correlation between the long-range regions. Liu et al. \cite{liu2018image} and Yu et al. \cite{yu2019free} focus on repairing the irregular shape of image damage, by capturing the spatial deformations of the damaged regions.
%
%
In addition to the semantic information, some works employ the geometric information like the edge and contour of image regions for effective image inpainting \cite{nazeri2019edgeconnect,ren2019structureflow}. 
%
%
%
However, the above methods generally formulate the inpainting task as the generation. Although the generated images look natural and realistic, they are less identical to the ground truth.

%
\begin{figure*}[t]
\centering
\includegraphics[width=1.0\linewidth]{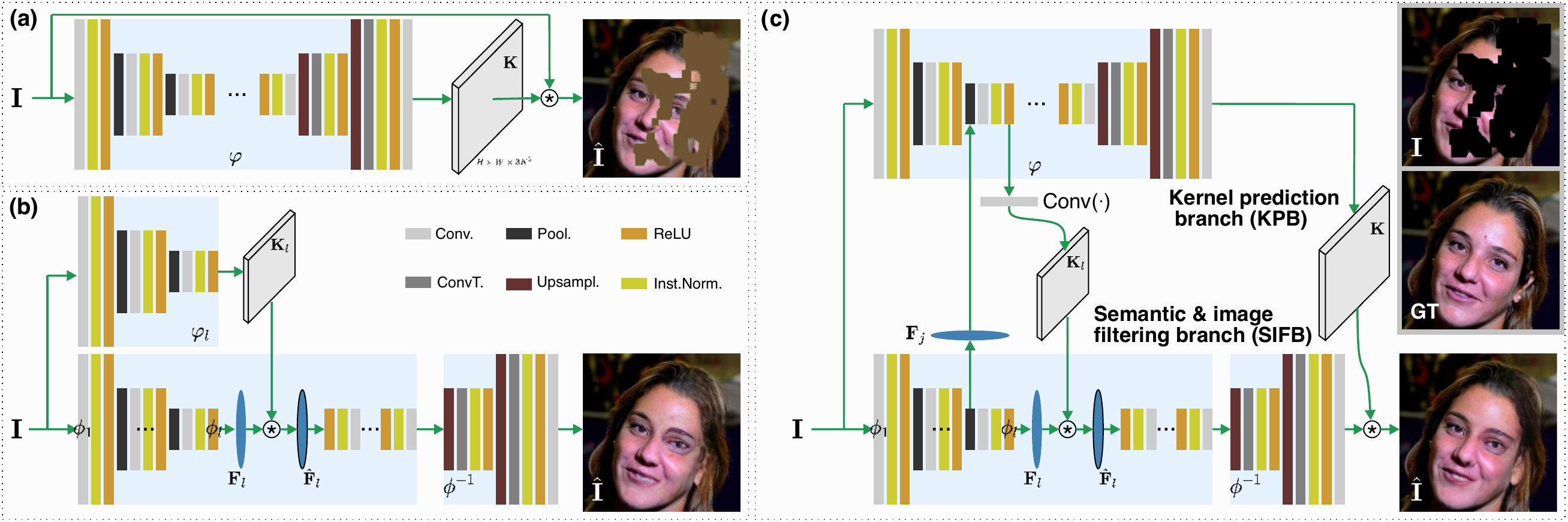}
\caption{
Three filtering-based image inpainting methods. (a) represents the predictive image-level filtering introduced in \secref{subsec:predfilt}. (b) shows the proposed semantic filtering-based inpainting in \secref{subse:semanticfilt}. (c) is the multi-level interactive siamese filtering(MISF) in \secref{subsec:misf}. We detail architectures in \tableref{tab:arch}.
}
\label{fig:frameworks}
\vspace{-5pt}
\end{figure*}
%

\highlight{Predictive filtering based image restoration}
%
%
The predictive filtering has been widely used in image restoration tasks, \eg,  denoising \cite{BakoACMTOG2017,mildenhall2018burst}, deraining \cite{guo2020efficientderain}, shadow removing \cite{fu2021auto}, and blur synthesis \cite{BrooksCVPR2019}. The predictive filtering allows more focused learning of the surrounding information for each pixel.
However, the image-level filtering can hardly address the tasks relying on the semantic understanding of the scene. 
Different from previous methods, for the first attempt, we propose a novel filtering method by extending the image-level filtering to the deep feature level. As a result, we can conduct semantic filtering and realize effective inpainting. Moreover, to take both advantages of image-level and deep feature-level filtering, we propose the multi-level semantic~\&~image filtering, which completes the semantic information as well as rich details.




\section{Discussion and Motivation}

\subsection{Predictive Filtering for Image Inpainting}
\label{subsec:predfilt}
Predictive filtering is a widely used image restoration technique and can address image denoising \cite{mildenhall2018burst} and deraining \cite{guo2020efficientderain} tasks. Here, we formulate the image inpainting as the pixel-wise predictive filtering task
%
\begin{align}\label{eq:pixelfilter-1}
\hat{\mathbf{I}} = \mathbf{I}\circledast\mathbf{K} ,
\end{align}
%
where $\mathbf{I}\in\mathds{R}^{H\times W}$ is the corrupted image and $\hat{\mathbf{I}}\in\mathds{R}^{H\times W}$ is the completed counterpart. The tensor $\mathbf{K}\in\mathds{R}^{H\times W\times N^2}$ contains $HW$ kernels for filtering all pixels. The operation `$\circledast$' denotes the pixel-wise filtering. We can expand the above equation as
%
\begin{align}\label{eq:pixelfilter-2}
\hat{\mathbf{I}}[\mathbf{p}] = \sum_{\mathbf{q}\in\mathcal{N}_{p}}\mathbf{K}_{p}[\mathbf{q}-\mathbf{p}]\mathbf{I}[\mathbf{q}].
\end{align}
%
Here, $\mathbf{p}$ and $\mathbf{q}$ are the coordinates of pixels in the image while the set $\mathcal{N}_p$ contains $N^2$ neighboring pixels of $\mathbf{p}$. The matrix $\mathbf{K}_p \in\mathds{R}^{N\times N}$ is the $p$th vector of $\mathbf{K}$ and determines the weights for all pixels in $\mathcal{N}_p$, which is also known as the kernel for the pixel $\mathbf{p}$.
Intuitively, the filtering is to reconstruct the pixel $\mathbf{p}$ by linearly combining its neighboring pixels. 
For image inpainting, the pixels at the boundary of missing areas are reasoned by their neighboring pixels. The principle is that the missing pixels do not break the local structure. Meanwhile, the related pixels can be used to reconstruct the missing pixels. 
However, the local structures around missing pixels are diverse and may distinguish them from each other.
To adapt the context variations, we can train a predictive network to estimate the kernels for all pixels according to the input image
%
\begin{align}\label{eq:kernel_predict}
\mathbf{K} = \varphi(\mathbf{I}),
\end{align}
%
where $\varphi(\cdot)$ is the predictive network. We set it as an encoder-decoder network (See \figref{fig:frameworks} (a)) and train it via the image quality loss (\ie, $L_1$) \cite{guo2020efficientderain}, GAN loss \cite{nazeri2019edgeconnect}, Style loss \cite{sajjadi2017enhancenet}, and the perceptual loss \cite{johnson2016perceptual}. We will detail these loss functions in \secref{subsec:impl}. The pipeline is shown in \figref{fig:frameworks}~(a). 

%
\begin{SCtable*}
\setlength{\tabcolsep}{4pt}
\footnotesize
\centering
\caption{Architecture of MISF. Network architectures of the encoder-decoder network (\ie, $\phi(\cdot)$ and$\phi^{-1}(\cdot)$) and the predictive network (\ie, $\varphi(\cdot)$). The variable $N$ is the kernel size of filtering, \ie, $|\mathcal{N}_p|=N^2$, and $C=3$ denotes the number of color channel. `conv(x,x,x)' defines the kernel size, input and output channel numbers, respectively.}\label{tab:arch}
\begin{tabular}{c|c|l|l||c|c|l|l}
\toprule
\multicolumn{4}{c||}{Encoder-decoder ($\phi^{-1}(\phi(\cdot))$)}                & \multicolumn{4}{c}{Predictive network ($\varphi(\cdot)$)}                     \\
\midrule
 In.           & Out.           & Out. size         & Layers                  & In.            & Out.           & Out. size         & Layers                    \\
\midrule
$\mathbf{I}$ & $\mathbf{F}_1$ & $256\times256$    & conv{(}7, 3, 64{)}       & $\mathbf{I}$ & $\mathbf{E}_1$ & $256\times256$    & conv{(} 7, 3, 64{)} \\
$\mathbf{F}_1$ & $\mathbf{F}_2$ & $128\times128$    & conv{(}4, 64, 128{)}     & $\mathbf{E}_1$ & $\mathbf{E}_2$ & $128\times128$    & conv{(} 4, 64, 128{)} \\
$\mathbf{F}_2$ & $\mathbf{F}'_2$ & $64\times64$      & \text{AvgPool}   & $\mathbf{E}_2$ & $\mathbf{E}'_2$ & $64\times64$    & \text{AvgPool} \\
$\mathbf{F}'_2$ & $\mathbf{F}_3$ & $64\times64$      & conv{(}4, 128, 256{)}    & $[\mathbf{F}'_2,\mathbf{E}'_2]$ & $\mathbf{E}_3$ & $64\times64$    & conv{(} 4, 256, 256{)} \\
$\mathbf{F}_3$ & $\hat{\mathbf{F}}_3$ & $64\times64$    & $\mathbf{F}_3\circledast\mathbf{K}_3$  & $\mathbf{E}_3$ & $\mathbf{K}_3$ & $64^2 \times 256 N^2$ & Conv{(}1, 256, $256 N^2${)} \\
$\hat{\mathbf{F}}_3$ & $\mathbf{F}_4$ & $64\times64$    & $8\times$ conv{(}1, 256, 256{)}  & $\mathbf{E}_3$ & $\mathbf{E}_4$ & $64\times64$    & $8\times$~conv{(} 1, 256, 256{)} \\
$\mathbf{F}_4$ & $\mathbf{F}_5$ & $64\times64$      & convt{(}4, 256, 128{)}   & $\mathbf{E}_4$ & $\mathbf{E}_5$ & $64\times64$  & convt{(} 4, 256, 128{)} \\
$\mathbf{F}_5$ & $\mathbf{F}_6$ & $128\times128$      & convt{(}4, 128, 64{)}    & $\mathbf{E}_5$ & $\mathbf{E}_6$ & $128\times128$    & convt{(} 4, 128, 64{)} \\
$\mathbf{F}_6$ & $\mathbf{F}_7$ & $256\times256$      & convt{(} 7, 64, $C${)}     & $\mathbf{E}_6$ & $\mathbf{K}$ & $256\times256$    & convt{(} 7, 64, $CN^2${)} \\
$\mathbf{F}_7$ & $\hat{\mathbf{I}}$ & $256\times256$ & $\mathbf{F}_7\circledast\mathbf{K}$  & - & - & - & - \\

\bottomrule
\end{tabular}
\vspace{-10pt}
\end{SCtable*}

%

%
\begin{figure}[t]
\centering
\includegraphics[width=1.0\linewidth]{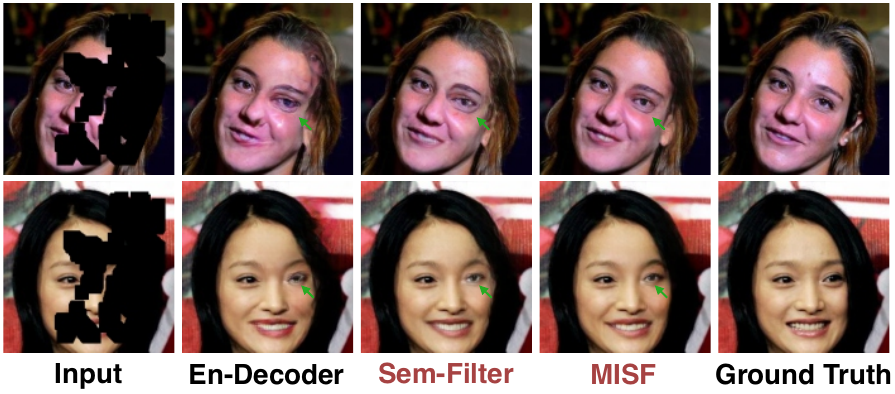}
\caption{
Two examples of the encoder-decoder network (En-Decoder), semantic filtering (Sem-Filter), and multi-level interactive siamese filtering
(MISF). We highlight the main differences via green arrows.
}
\label{fig:obs2}
\vspace{-10pt}
\end{figure}
%

\subsection{Challenges and Motivations}
\label{subsec:challenges}
The above predictive filtering for inpainting is non-trivial and should be carefully studied. 
Specifically, we can train the function $\varphi(\cdot)$ with an image inpainting dataset like CelebA dataset \cite{liu2015deep}. Then, we evaluate it on a series of images with the missing areas becoming larger and thicker. 
We show an example in \figref{fig:observation} and observe that: \ding{182} for the thin and small missing areas, the predictive filtering can complete the missing pixels effectively and lead to high-fidelity results (See \figref{fig:observation} (a) and (d)). Nevertheless, when the missing areas become larger and thicker, the pixels away from the missing areas' boundaries cannot be recovered. This is because the large missing pixels break the local structure. Thus, the image-level filtering cannot achieve the reconstruction goal anymore.
\ding{183} Different scenes require the predicted kernels to adapt the semantic variations. Nevertheless, the image-level filtering can only reconstruct pixels according to their local contexts and cannot understand the whole scene. For example, when the missing area is significantly large (See \figref{fig:observation} (c)), the image-level filtering cannot guess what pixels should be filled to make the face realistic with high fidelity.

A naive solution for the challenges is to conduct the filtering recurrently. Specifically, we can perform filtering on the inpainting result, again and again, that is, we use the estimated missing pixels to reconstruct the pixels inside the missing areas.
We show the result of such a strategy in \figref{fig:observation} (h) for inpainting (c). The completed pixels become vague around the center of the missing areas. 
It is mainly because the large missing areas break the local structure. Thus, only the pixels near the boundary are reconstructed but have low fidelity. The reconstruction errors are accumulated during the recurrent filtering process.
Recently, Guo \etal\cite{guo2021jpgnet} incorporate the predictive filtering and generative network to address this issue. However, such a solution can equitably introduce some artifacts of the state-of-the-art generative network-based method.
As a result, a novel technique is necessary to address the challenges.

\section{Methodology}

\subsection{Semantic Filtering for Image Inpainting}
\label{subse:semanticfilt}
As explained in the \secref{subsec:challenges}, the image filtering-based inpainting is not that effective since the large missing areas break the local structure information that lays the foundation of filtering-based restoration. 
To address this issue, we propose to extend the filtering from the image level to the deep feature level that contains semantic information. 
The intuitive idea is that semantic information can be preserved even a large area of the image is lost.
As the case in \figref{fig:observation} (c), even though the large areas of the girl's face are missed and a human can fill the missing regions according to the understanding of the face.
To achieve the semantic filtering, we first employ an encoder-decoder network where the encoder is to extract features of the corrupted image (\ie, $\mathbf{I}$) and the decoder is to map the features to the completed image. We have the following formulation for the encoder
%
\begin{align}\label{eq:encoder}
\mathbf{F}_L = \phi(\mathbf{I}) = \phi_L(\ldots\phi_l(\ldots\phi_2(\phi_1(\mathbf{I}))))
\end{align}
%
where $\phi(\cdot)$ is the encoder and $\mathbf{F}_l$ is the deep feature extracted from the $l$th layer, \ie, $\mathbf{F}_l=\phi_l(\mathbf{F}_{l-1})$. For example, $\mathbf{F}_L$ is the output of the last layer of $\phi(\cdot)$ (\ie, $\phi_L(\cdot)$). The decoder can be formulated as 
%
\begin{align}\label{eq:decoder}
\hat{\mathbf{I}} = \phi^{-1}(\mathbf{F}_L),
\end{align}
%
where $\phi^{-1}(\cdot)$ is the decoder.
Then, we conduct the semantic filtering on extracted features like the image-level filtering
%
\begin{align}\label{eq:semanticfilter}
\hat{\mathbf{F}}_l[\mathbf{p}] = \sum_{\mathbf{q}\in\mathcal{N}_{p}}\mathbf{K}_{p}^l[\mathbf{q}-\mathbf{p}]\mathbf{F}_l[\mathbf{q}],
\end{align}
%
where $\mathbf{K}_{p}^l$ is the kernel for filtering the $p$th element of $\mathbf{F}_l$ via the neighboring elements, \ie, $\mathcal{N}_p$. 
We use the matrix $\mathbf{K}_l$ to include all element-wise kernels (\ie, $\mathbf{K}_{p}^l$).
After that, we replace the $\mathbf{F}_l$ with $\hat{\mathbf{F}}_l$ in \reqref{eq:encoder} and conduct the subsequent operations. 
To let the kernels adapt to different scenes, we also employ a predictive network to predict the kernels like the image-level predictive filtering (\ie, \reqref{eq:kernel_predict})
%
\begin{align}\label{eq:featkernel_predict}
\mathbf{K}_l = \varphi_l(\mathbf{I}),
\end{align}
%
where $\varphi_l(\cdot)$ is the predictive network to produce $\mathbf{K}_l$.

%
\begin{figure*}
\centering
\includegraphics[width=0.97\linewidth]{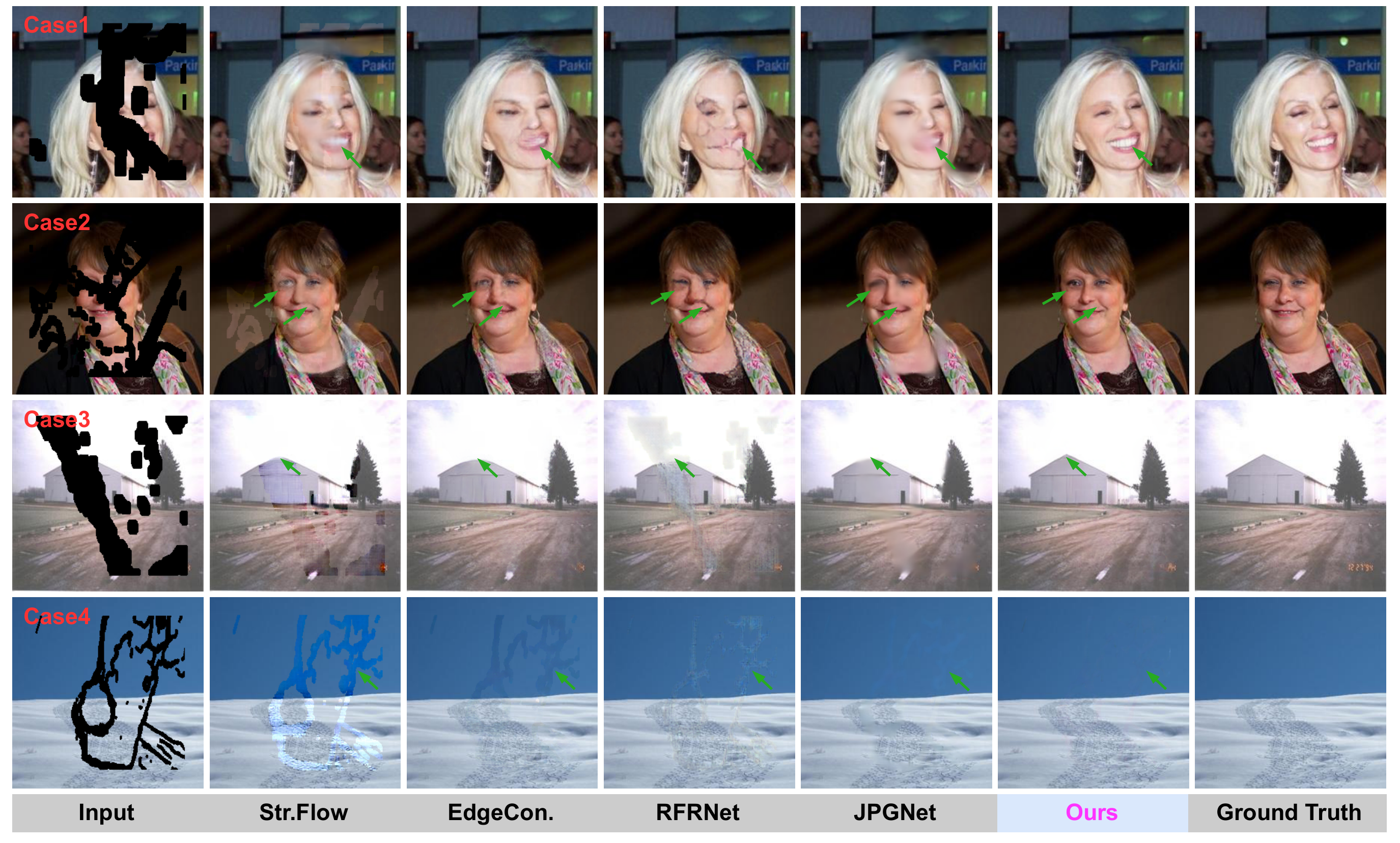}
\vspace{-5pt}
\caption{ Four visualization results of Str.Flow\cite{ren2019structureflow}, EdgeCon.\cite{nazeri2019edgeconnect}, RFRNet\cite{li2020recurrenta}, JPGNet\cite{guo2021jpgnet}, and ours method. The case1 and case2 are from the CelebA dataset, the case3 and case4 are from Places2 dataset. We highlight the main differences via green arrows.
}
\label{fig:exp1}
\vspace{-10pt}
\end{figure*}

\begin{figure}[t]
\centering
\includegraphics[width=1.0\linewidth]{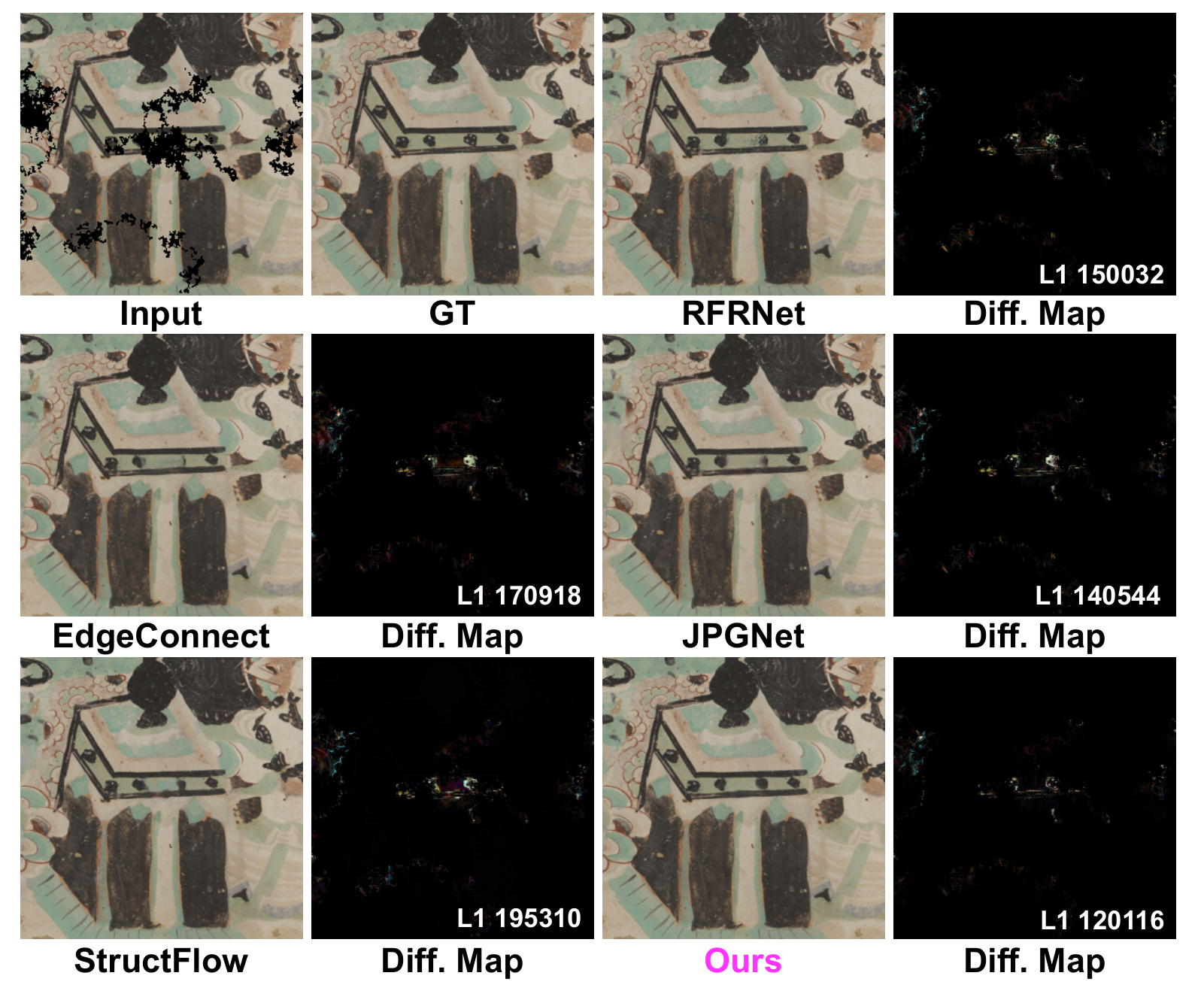}
\vspace{-10pt}
\caption{An example from the Dunhuang dataset. In addition to the inpainting results of five methods, we show the difference maps and L1 norm between the predicted results and the ground truth (GT). }\label{fig:dunhuang}
\vspace{-15pt}
\end{figure}

%
We show the semantic filtering-based image inpainting in \figref{fig:frameworks} (b) and train the networks (\ie, $\phi(\cdot)$, $\phi^{-1}(\cdot)$, and $\varphi_l(\cdot)$) via the $L_1$, GAN, Style and perceptual loss functions like the predictive filtering.
We present the inpaiting examples in \figref{fig:frameworks} and \figref{fig:obs2} and have the following observations: \ding{182} 
Compared with the image-level predictive filtering, semantic filtering can fill all missing pixels and recovery the semantic information effectively. As the cases in \figref{fig:obs2}, the structures of the missed left eyes and faces are recovered. As a result, the inpainting results are more realistic and have higher fidelity. \ding{183} Although the main structures are recovered, the results lose details. In the first case of \figref{fig:obs2}, the girl's forehead and left eye still contain artifacts and her mouth is blurred. We have similar observations on other cases.
%

\subsection{Multi-level Interactive Siamese Filtering}
\label{subsec:misf}
Semantic filtering fills the missing semantic information at the deep feature level that has a low spatial resolution. Thus, it inevitably loses detailed information. 
To solve this problem, a straightforward solution is to conduct filtering on multi-level features. 
For example, for all features extracted from the encoder (\ie, $\{\mathbf{F}\}_{l=1}^L$), we can filter each of them via an exclusive predictive network like \secref{subse:semanticfilt}, which, however, would lead to extra memory and time costs and cannot take the advantages of features from different layers.
To address this issue, we propose the \textit{multi-level interactive Siamese filtering (MISF)} that consists of two branches with similar architectures, \ie, \textit{kernel prediction branch (KPB)} and \textit{semantic~\&~image filtering branch (SIFB)}, which are encoder-decoder networks containing several convolutional blocks. 
The two branches are interactively linked: KPB (\ie, $\varphi(\cdot)$ in \figref{fig:frameworks} (c)) takes the raw image and multiple features of SIFB as inputs and predicts multi-level kernels for SIFB. SIFB  (\ie, $\phi(\cdot)$ in \figref{fig:frameworks} (c)) uses these kernels to filter the features at different levels. As a result, SIFB are dynamically changed according to the input. We show the whole framework in \figref{fig:frameworks} (c).

Specifically, given a corrupted image $\mathbf{I}$, we feed it to the SIFB that conducts filtering at the image level and semantic level (\ie, filtering at the $l$th-layer feature) jointly. As a result, we can generate the completed image by 
%
\begin{align}\label{eq:mdl_filtering}
\hat{\mathbf{I}} = \phi^{-1}(\phi_L(\ldots\phi_{l+1}(\mathbf{F}_l\circledast\mathbf{K}_l)))\circledast\mathbf{K},
\end{align}
%
where $\mathbf{F}_l = \phi_l(\ldots\phi_1(\mathbf{I}))$. The kernels for deep feature and image (\ie, $\mathbf{K}_l$ and $\mathbf{K}$) are predicted by the KPB
%
\begin{align}
& \mathbf{K}_l = \text{Conv}(\varphi_{l}(\ldots\varphi_{j+1}([\mathbf{E}_{j},\mathbf{F}_{j}]))), \label{eq:kpb-1}\\
& \mathbf{K} = \varphi_{L}(\ldots\varphi_{j+1}([\mathbf{E}_{j},\mathbf{F}_{j}])), \label{eq:kpb-2}
\end{align}
%
where $\mathbf{F}_j = \phi_j(\ldots\phi_1(\mathbf{I}))$ is the features from the $j$th layer of SIFB, and $\mathbf{E}_j = \varphi_j(\ldots\varphi_1(\mathbf{I}))$ is from the $j$th layer of KPB. We add a convolutional layer ($\text{Conv}(\cdot)$) to adjust the size of $\varphi_{l}(\ldots\varphi_{j+1}([\mathbf{E}_{j},\mathbf{F}_{j}]))$ to meet the requirements of kernels. We show the whole framework in \figref{fig:frameworks} (c). 
The kernels $\mathbf{K}_l$ and $\mathbf{K}$ are for the feature-level and image-level filtering, respectively.
Intuitively, with \reqref{eq:mdl_filtering}, we conduct both semantic~\&~image filtering in a single framework to fill large missing regions and enhance details.
Moreover, with \reqref{eq:kpb-1} and \reqref{eq:kpb-2}, all predicted kernels for semantic~\&~image filtering are driven by the input image $\mathbf{I}$ and the deep feature $\mathbf{F}_j$, which contain all available spatial details and the understanding of the whole scene. As a result, both semantic information and detailed pixels can be properly reconstructed.
With the new designs, our method achieves high-fidelity image inpainting. As shown in \figref{fig:obs2}, MISF produces semantic face structures with rich details.

\subsection{Relationship to Encoder-Decoder Network}
In this section, we aim to explain the effectiveness of our method from the viewpoint of the encoder-decoder network. 
We can use the naive encoder-decoder network to perform the image inpainting directly.
For example, we feed the corrupted image into an encoder and use a decoder to reconstruct the image.
This process can be represented through \reqref{eq:encoder} and \reqref{eq:decoder}. 
We can train the encoder-decoder network via the same loss functions like filtering. 

From the perspective of the encoder-decoder network, our semantic filtering is an improved encoder-decoder network that contains an extra `dynamic convolution layer' (See \figref{fig:frameworks} (b)). MISF further makes the dynamic process conditional on the multi-level features. As a result, the parameters of the dynamic convolution are element-wise and dynamically tuned according to different images and their semantic meaning through the predictive network.
The advantages of dynamic convolution have been evidenced in many works \cite{chen2020dynamic,guo2021sparta}.
However, these works mainly focus on the image classification task. They predict convolutional parameters dynamically, according to the input features. 
In contrast, our work presents the importance of dynamic convolution for image inpainting and predicts dynamic convolutional parameters based on the raw input and deep features jointly with an element-wise way.
According to the results in \figref{fig:obs2}, we see that the proposed dynamic operation is critical to the high-quality inpainting results. 
With the same training setups (See \secref{subsec:impl}), the naive encoder-decoder network produces artifacts on the missing areas. %
The filled pixels lead to obvious structure mismatches, while semantic filtering can adapt different scenes and fill pixels to have reasonable structures.
Moreover, the complete model of MISF, which considers both semantic~\&~image filtering,  achieves much better results of semantic and detail recovery.

\subsection{Implementation Details}
\label{subsec:impl}
%

\highlight{Network architectures} We detail the architectures in \tableref{tab:arch}. Theoretically, we conduct the MISF on all deep features. Nevertheless, this will lead to significant memory and time costs. Here, we only perform semantic filtering at the $3$th layer (\ie, $\mathbf{F}_3$). We will discuss semantic filtering on other features in the experiment section.

\begin{SCtable*}
\footnotesize
\centering
\caption{Comparison results on Places2, CelebA, and Dunhuang datasets. For PConv, the reported result is from the paper \cite{liu2018image}}
\small
{
    \resizebox{1.6\linewidth}{!}{
    {
	\begin{tabular}{l|l|ccc|ccc|c}
		
    \toprule
     & \multicolumn{1}{c|}{Datasets} & \multicolumn{3}{c|}{Places2} & \multicolumn{3}{c|}{CelebA}  & \multicolumn{1}{c}{Dunhuang} \\
    & \multicolumn{1}{c|}{Mask Ratio} &  0\%-20\% & 20\%-40\% & 40\%-60\% & 0\%-20\% & 20\%-40\% & 40\%-60\% & Default\\
    \midrule
    \multirow{6}{*}{PSNR $\uparrow$}
    
    & PConv \cite{liu2018image} & 31.030  & 23.673  & 19.743 & -   & -  & - & -\\
    
    & StructFlow \cite{ren2019structureflow} & 29.047 & 23.092 & 19.408 & 31.618 & 25.283 & 20.829 & 35.199  \\
    
    & EdgeConnect \cite{nazeri2019edgeconnect} & 29.899 & 23.378 & 19.522 & 32.781 & 25.347 & 20.449 & 36.419 \\
    
    & RFRNet \cite{li2020recurrenta} & 29.281 & 22.589 & 18.581 & 33.573 & 25.635 & 20.539 & 36.485\\
    
    & JPGNet \cite{guo2021jpgnet} & 30.673 & 23.937 & 19.884 & 34.401 & 26.543 & 21.297 & 37.646\\
    
    & CTSDG \cite{guo2021image}  & 30.658 & 23.701 & 19.751 & 32.677 & 24.945 & 20.123 & - \\
   
     &  \topone{MISF} &  \topone{31.335} &  \topone{24.239}  &  \topone{20.044}  &  \topone{34.494}   &  \topone{26.635}  &  \topone{21.553} &  \topone{38.383} \\
     						
    \midrule
    \multirow{6}{*}{SSIM $\uparrow$}  
    
    & PConv \cite{liu2018image} & 0.9070  & 0.7310  & 0.5325 & -   & -  & - & -\\
    
    & StructFlow \cite{ren2019structureflow} & 0.9343 & 0.8187 & 0.6740 & 0.9487 & 0.8598 & 0.7417 & 0.9559  \\
    
    & EdgeConnect \cite{nazeri2019edgeconnect} & 0.9396 & 0.8225 & 0.6710 & 0.9586 & 0.8689 & 0.7362 & 0.9635 \\

    & RFRNet \cite{li2020recurrenta} & 0.9283 & 0.7868 & 0.6137 & 0.9626 & 0.8746 & 0.7400 & 0.9648\\
    
    & JPGNet \cite{guo2021jpgnet} & 0.9452 & 0.8348 & 0.6915 & 0.9674 & 0.8908 & 0.7697 & 0.9724\\
    
    & CTSDG \cite{guo2021image}  & 0.9451 & 0.8299 & 0.6768 & 0.9587 & 0.8649 & 0.7291 & - \\

    &  \topone{MISF} &  \topone{0.9506} &  \topone{0.8435}  &  \topone{0.6931}  &  \topone{0.9680}   &  \topone{0.8911}  &  \topone{0.7698} &  \topone{0.9735} \\

    \midrule
    \multirow{6}{*}{L1 $\downarrow$}  
    & PConv \cite{liu2018image} & 0.808  & 2.495  & 5.098 & -   & -  & - & -\\
    
    & StructFlow \cite{ren2019structureflow} & 0.976 & 2.811 & 5.444 & 0.737 & 2.171 & 4.533 & 0.475  \\
    
    & EdgeConnect \cite{nazeri2019edgeconnect} & 0.848 & 2.606 & 5.302 & 0.579 & 1.922 & 4.485 & 0.441 \\
    
    & RFRNet \cite{li2020recurrenta} & 1.009 & 3.218 & 6.719 & 0.521 & 1.811 & 4.346 & 0.401\\
    
    & JPGNet \cite{guo2021jpgnet} & 0.830 & 2.581 & 5.294 & 0.477 & 1.651 & 4.042 & 0.353\\
    
    & CTSDG \cite{guo2021image}  & 1.568 & 4.987 & 10.29 & 1.161 & 3.972 & 9.231 & - \\
    
    &  \topone{MISF}  &  \topone{0.726} &  \topone{2.340}  &  \topone{4.965}  &  \topone{0.474}   &  \topone{1.616}  &  \topone{3.826} &  \topone{0.341} \\

    \midrule
    \multirow{6}{*}{LPIPS $\downarrow$}  
    & PConv \cite{liu2018image} & -  & -  & - & -   & -  & - & -\\
    
    & StructFlow \cite{ren2019structureflow} & 0.0716 & 0.1714 & 0.2845 & 0.0746 & 0.1683 & 0.2662 & 0.0589  \\
    
    & EdgeConnect \cite{nazeri2019edgeconnect} & 0.0572 & 0.1541 & 0.2748 & 0.0456 & 0.1265 & 0.2380 & 0.0480 \\
    
    & RFRNet \cite{li2020recurrenta} & 0.0825 & 0.2161 & 0.3571 & 0.0400 & 0.1215 & 0.2335 & 0.0463\\
    
    & JPGNet \cite{guo2021jpgnet} & 0.0817 & 0.2145 & 0.3535 & 0.0440 & 0.1316 & 0.2502 & 0.0469\\
    
    & CTSDG \cite{guo2021image}  & 0.0525 & 0.1522 & 0.2714 & 0.0417 & 0.1267 & 0.2377 & - \\
    
    &  \topone{MISF}  &  \topone{0.0432} &  \topone{0.1298}  &  \topone{0.2499}  &  \topone{0.0315}   &  \topone{0.0949}  &  \topone{0.1911} &  \topone{0.0330} \\
    
    \bottomrule

	\end{tabular}
	}
	}
}
\label{tab:comparison}
\vspace{-10pt}
\end{SCtable*}

\highlight{Loss functions} To get high-fidelity images in both image quality and semantic levels, we follow the work \cite{nazeri2019edgeconnect} and train the networks with four loss functions, \ie, $L_1$ loss, GAN loss, Style loss, and perceptual loss.
Specifically, given a corrupted image $\mathbf{I}$, the predicted completion $\hat{\mathbf{I}}$, and the ground truth $\mathbf{I}^*$, we have the loss function
%
\begin{align}\label{eq:loss}
\mathcal{L}(\hat{\mathbf{I}}, \mathbf{I}^*) = \lambda_1\mathcal{L}_{1}+\lambda_2\mathcal{L}_{\text{gan}}+\lambda_3\mathcal{L}_{\text{perc}}+\lambda_4\mathcal{L}_{\text{style}}.
\end{align}
%
We fix $\lambda_1=1$, $\lambda_2=\lambda_3=0.1$, and $\lambda_4=250$. Please find the definitions of the loss functions in \cite{nazeri2019edgeconnect}.

\highlight{Training details} For all variants of our method, we use the same training setup: we employ Adam as the optimizer with the learning rate of 0.0001. We train the network for about 350,000 iterations with a batch size of 16. The experiments are implemented on the same platform with two NVIDIA Tesla V100 GPUs.

\section{Experiments}
\subsection{Setups}


{\bf Datasets.} We evaluate our method on three datasets, \ie, Places2 challenge dataset \cite{zhou2017places}, CelebA dataset \cite{liu2015deep}, and Dunhuang Challenge \cite{yu2019dunhuang}. The Places2 dataset contains over eight million images captured under over 365 scenes. The CelebA dataset contains over 180 thousand face images. These datasets allow our method to be evaluated on the natural and facial scenes. The Dunhuang challenge provides practical data for image inpainting. We evaluate the proposed method on the standard test set of the CelebA and Dunhuang Challenge datasets. For the Places2 dataset, we follow the convention and choose 30,000 random images for testing.


{\bf Metrics.} We follow the common setups in the image inpainting. We use the peak signal-to-noise ratio (PSNR), structural similarity index (SSIM), $L_1$, and perceptual similarity (LPIPS \cite{zhang2018unreasonable}) for measuring the quality of image inpainting.
%
%
%
PSNR, SSIM, and $L_1$ measures the quality of the recovered image. LPIPS measures the perceptual consistence between the recovered images and ground truth.

%


{\bf Mask Setups.} For the Places2 and CelebA datasets, we use the irregular mask dataset \cite{liu2018image}, which has been used in many works \cite{ren2019structureflow}, to generate the corrupted images. The mask images are classified into three categories (\ie, $0\%-20\%$, $20\%-40\%$, and $40\%-60\%$), based on the proportion of the image occupied by the holes. For the Dunhuang dataset, we follow its official setup.


{\bf Baselines.} We compare with five state-of-the-art inpainting methods, including PConv \cite{liu2018image}, StructFlow \cite{ren2019structureflow}, Edge-Connect \cite{nazeri2019edgeconnect}, RFR-Net \cite{li2020recurrenta}, JPGNet \cite{guo2021jpgnet}, and CTSDG\cite{guo2021image}.

\begin{table*}
\centering
\caption{Ablation study results on Places2, CelebA, and Dunhuang datasets.}
\small
{
    \resizebox{0.8\linewidth}{!}{
    {
	\begin{tabular}{l|l|ccc|ccc|c}
		
    \toprule
     & \multicolumn{1}{c|}{Datasets} & \multicolumn{3}{c|}{Places2} & \multicolumn{3}{c|}{CelebA}  & \multicolumn{1}{c}{Dunhuang} \\
    & \multicolumn{1}{c|}{Mask Ratio} &  0\%-20\% & 20\%-40\% & 40\%-60\% & 0\%-20\% & 20\%-40\% & 40\%-60\% & Default\\
    \midrule
    \multirow{6}{*}{PSNR $\uparrow$}  
    & Img-Filter  & 25.489 & 17.663 & 13.773 & 27.314 & 19.020 & 14.837 & 37.021 \\
     
    & Sem-Filter & 31.010 & 24.117 & 19.944 & 34.253 & 26.518 & 21.486 & 37.897\\
    
    & \topone{MISF} &  \topone{31.335} &  \topone{24.239}  &  \topone{20.044}  &  \topone{34.494}   &  \topone{26.635}  &  \topone{21.553} &  \topone{38.383} \\
    
     & En-decoder-Filter & 31.187 & 24.107 & 19.898 & 34.330 & 26.484 & 21.428 & 38.195 \\
    
    & En-decoder & 30.824 & 23.980 & 19.871 & 33.745 & 26.122 & 21.077 & 37.766  \\
    
    \midrule
    \multirow{6}{*}{SSIM $\uparrow$}  
    & Img-Filter  & 0.9180 & 0.7571 & 0.5911 & 0.9298 & 0.7830 & 0.6301 & 0.9692 \\
    
    & Sem-Filter & 0.9487 & 0.8409 & 0.6910 & 0.9657 & 0.8871 & 0.7631 & 0.9696\\
    
    & \topone{MISF} &  \topone{0.9506} &  \topone{0.8435}  &  \topone{0.6931}  &  \topone{0.9673}   &  \topone{0.8909}  &  \topone{0.7693} &  \topone{0.9735} \\
    
    & En-decoder-Filter & 0.9499 & 0.8420 & 0.6907 & 0.9661 & 0.8871 & 0.7614 & 0.9734 \\
    
    & En-decoder & 0.9466 & 0.8358 & 0.6841 & 0.9632 & 0.8805 & 0.7510 & 0.9711  \\

    \midrule
    \multirow{6}{*}{L1 $\downarrow$}  
    & Img-Filter  & 1.535 & 5.477 & 11.42 & 1.264 & 4.750 & 10.44 & 0.386 \\
    
    & Sem-Filter & 0.750 & 2.370 & 4.995 & 0.488 & 1.651 & 3.895 & 0.383\\
    
    & \topone{MISF}  &  \topone{0.726} &  \topone{2.340}  &  \topone{4.965}  &  \topone{0.474}   &  \topone{1.616}  &  \topone{3.826} &  \topone{0.341} \\
    
    & En-decoder-Filter & 0.732 & 2.360 & 5.022 & 0.487 & 1.657 & 3.909 & 0.345 \\
    
    & En-decoder & 0.769 & 2.432 & 5.104 & 0.518 & 1.740 & 4.117 & 0.362  \\

    \midrule
    \multirow{6}{*}{LPIPS $\downarrow$}  
    
    & Img-Filter  & 0.1154 & 0.3129 & 0.5168 & 0.1084 & 0.2910 & 0.4631 & 0.0508 \\
    
    & Sem-Filter & 0.0465 & 0.1333 & 0.2527 & 0.0343 & 0.1008 & 0.2009 & 0.0379\\
    
    & \topone{MISF}  &  \topone{0.0432} &  \topone{0.1298}  &  \topone{0.2499}  &  \topone{0.0315}   &  \topone{0.0949}  &  \topone{0.1911} &  \topone{0.0330} \\
    
    & En-decoder-Filter & 0.0444 & 0.1328 & 0.2540 & 0.0346 & 0.1016 & 0.2015 & 0.0332 \\
    
    & En-decoder & 0.0497 & 0.1416 & 0.2637 & 0.0370 & 0.1066 & 0.2104 & 0.0373  \\
    \bottomrule
		
	\end{tabular}
	}
	}
}
\label{tab:ablation}
\vspace{-10pt}
\end{table*}

\subsection{Comparison Results}

{\bf Quantitative comparison.}
%
%
We compare our method with five state-of-the-art inpainting methods on three public datasets. As the results in \tableref{tab:comparison}, we have the following observations:
\ding{182} Our method achieves better PSNR, SSIM, and $L_1$ scores across all datasets and mask ratios than other competitive methods. Compared to RFRNet, we achieve 7.01\% relative higher PNSR under the $0\%-20\%$ mask ratio on the Places2 dataset. Moreover, the relative gaps become larger, \ie, 7.3\% and 7.9\%, under $20\%-40\%$ and $40\%-60\%$ mask ratios, respectively. It demonstrates that our method presents obvious advantages over existing methods on high-fidelity restoration.
%
%
\ding{183} When considering the LPIPS, we have similar observations. Our method gets 47.12\% relative lower LPIPS than JPGNet under the $0\%-20\%$ mask ratio on the Place2 dataset. This result demonstrates the impressive progress of our method on perceptual recovery.
\ding{184} The consistent advantages across different datasets and mask ratios demonstrate that our method has high generalization capability.




{\bf Qualitative comparison.}
%
%
We provide the visualization results on five cases taken from three datasets (\ie, CelebA, Place2, and Dunhuang) in \figref{fig:exp1} and \figref{fig:dunhuang}. We find that: \ding{182} Our method generates more natural and high-fidelity images that are significantly close to the ground truth, even the missing areas are large (See \figref{fig:exp1} case1). On the other hand, other methods introduce many artifacts like structure distortions and blur of large missing areas. \ding{183} Though all methods present similar results on small missing areas (\eg, \figref{fig:dunhuang}), our method provides fine-grained structures and recovers better details. For example, for the local structures around the green arrows in case3, RFRNet, EdgeCon., JPGNet and Str.~Flow fail to restore the detailed structures. In contrast, our method completes all the details properly.


\begin{table}
\centering
\setlength{\tabcolsep}{3pt}
\caption{Sem-Filter with different deep features.}
\small
{
    \resizebox{0.8\linewidth}{!}{
    {
	\begin{tabular}{l|l|ccc}
		
    \toprule
     & \multicolumn{1}{c|}{Datasets}  & \multicolumn{3}{c}{CelebA} \\
    & \multicolumn{1}{c|}{Mask Ratio} & 0\%-20\% & 20\%-40\% & 40\%-60\%\\
    \midrule
    \multirow{3}{*}{PSNR $\uparrow$}  
    & Sem-Filter($\mathbf{F}_1$)  & 33.981 & 26.364 & 21.360 \\
     
    & Sem-Filter($\mathbf{F}_2$) & 34.128 & 26.429 & 21.353 \\
    
    & Sem-Filter($\mathbf{F}_3$)  &  \topone{34.253} & \topone{26.518} & \topone{21.486}  \\
    
    \midrule
    \multirow{3}{*}{SSIM $\uparrow$}  
    & Sem-Filter($\mathbf{F}_1$)  & 0.9651 & 0.8856 & 0.7612\\
     
    & Sem-Filter($\mathbf{F}_2$) & 0.9651 & 0.8852 & 0.7589 \\
    
    & Sem-Filter($\mathbf{F}_3$) & \topone{0.9657} & \topone{0.8871} & \topone{0.7631}  \\

    \midrule
    \multirow{3}{*}{L1 $\downarrow$}  
    & Sem-Filter($\mathbf{F}_1$)  & 0.503 & 1.692 & 3.977 \\
     
    & Sem-Filter($\mathbf{F}_2$) & 0.497 & 1.673 & 3.952 \\
    
    & Sem-Filter($\mathbf{F}_3$)  &  \topone{0.488} & \topone{1.651} & \topone{3.895}  \\

    \midrule
    \multirow{3}{*}{LPIPS $\downarrow$}  
    
    & Sem-Filter($\mathbf{F}_1$)  & 0.0359 & 0.1038 & 0.2043\\
     
    & Sem-Filter($\mathbf{F}_2$) & 0.0355 & 0.1021 & 0.2028 \\
    
    & Sem-Filter($\mathbf{F}_3$)  &  \topone{0.0343} & \topone{0.1008} & \topone{0.2009}  \\
    
    \bottomrule
		
	\end{tabular}
	}
	}
}
\label{tab:discussion}
\vspace{-10pt}
\end{table}

\subsection{Ablation Study}

{\bf Quantitative results.} To validate the effectiveness of MISF, we consider three variants: Img-Filter (\secref{subsec:predfilt}), Sem-Filter (\secref{subse:semanticfilt}), and MISF \secref{subsec:misf}.
%
%
In \tableref{tab:ablation}, we observe that: \ding{182} By combining the image-level and semantic filtering, MISF yields better scores on all metrics across the three datasets and three missing sizes than other methods.
%
%
\ding{183} Img-Filter yields poor scores on the Places2 and CelebA datasets but has good results in the Dunhuang dataset. Note that the results of Img-Filter on Dunhuang are even better than RFRNet. It is mainly because Img-Filter is good at completing small missing holes but failing to address large missing areas, as analyzed in \secref{subsec:predfilt}.


{\bf Qualitative results.} 
%
%
We visualize the results of Img-Filter in \figref{fig:observation}. It shows that Img-Filter completes small missing areas effectively. However, Img-Filter is less effective in terms of handling large areas. We also compare Sem-Filter and MISF in \figref{fig:obs2}. It presents Sem-Filter loses many details while MISF produces rich details with natural structures.


\begin{figure*}[t]
\centering
\includegraphics[width=1.0\linewidth]{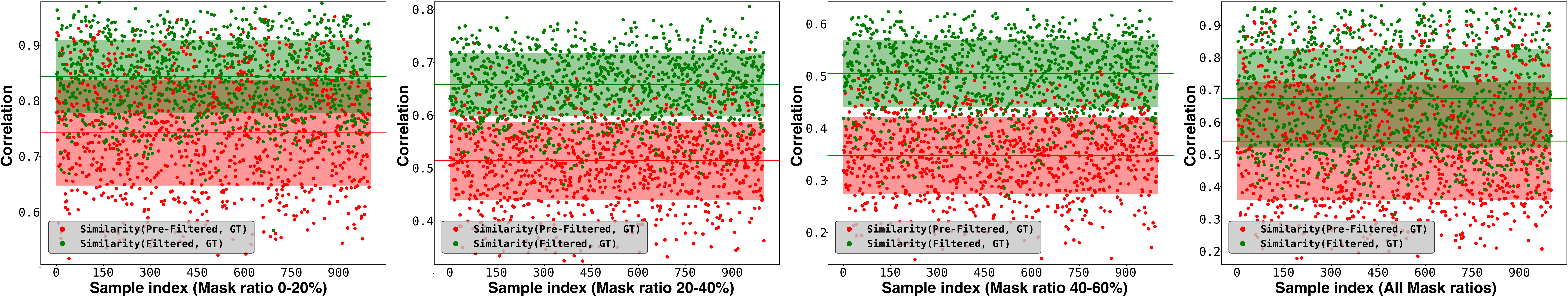}
\caption{Similarity of filtered and pre-filtered features to ground truth features.}\label{fig:rebuttal3}
\vspace{-10pt}
\end{figure*}


\subsection{Discussion}

\highlight{Filtered vs. Pre-Filtered Features.} Given a corrupted image and its ground truth (GT), we feed them to MISF and get their deep features before~\&~after filtering, respectively. Then, we calculate the similarity via cross-correlation between the filtered (or pre-filtered) features of the corrupted image and GT (See in Fig.~\ref{fig:rebuttal3}). We randomly sample 1000 examples from different mask ratios and the whole dataset, respectively, and calculate each example's similarity. After filtering, the features of corrupted image become close to the ones of GT. As the mask ratio is larger, the similarity margin becomes larger, which infers the effectiveness of semantic filtering across different mask sizes.

\highlight{Semantic filtering with different deep features}
%
As the results reported in \tableref{tab:discussion}, the completion performance generally becomes better when the feature for semantic filtering is deeper.
We have consistent conclusions on the four metrics and three missing sizes.
%
This is because deeper features have better semantic representations.
Nevertheless, whether the growth would continue as the network become deeper should be further studied in the future.


\highlight{Importance of dynamic convolutional operations for the encoder-decoder network}
%
Our method can be regarded as an advanced encoder-decoder network, which contains dynamic convolutional operations.
To validate this, we conduct an ablation study from the viewpoint encoder-decoder network.
Specifically, we compare four variants, \ie, En-decoder, En-decoder-Filter, Sem-Filter, and MISF.
%
The first one is the naive encoder-decoder network without the dynamic convolutional operation. The second variant is built by adding an image-level predictive filtering to the output of En-decoder. As a result, En-decoder-Filter and Sem-Filter can be regarded as the encoder-decoder network that contains a single dynamic convolutional operation.
MISF has two dynamic convolutional operations.
%
Comparing all variants, we see that the networks with more dynamic convolutional operations yield better inpainting results under the four metrics across all datasets and missing areas.


\section{Conclusions}

We proposed multi-level interactive siamese filtering (MISF) for high-fidelity image inpainting. We use a single predictive network to conduct predictive filtering at the image level and deep feature level, simultaneously. The predictive network takes the raw input image and deep features to predict the filtering kernels at different levels. As a result, the predicted kernels contain information for joint semantic and pixel filling.
Specifically, the image-level filtering is to recover details, while the deep feature-level filtering is to complete semantic information, which leads to high-fidelity inpainting results. In addition, the dynamically predicted kernels make our method have high generalization capability. Our method outperforms state-of-the-art methods on three public datasets. Furthermore, the extensive experiments demonstrate the effectiveness of different components of our approach. One potential limitation of this work is that we validate and train our model on the widely used public datasets that may cover a part of the real-world scenarios. In the future, we could develop our model to see more scenarios and further enhance its generalization capability.

\small
\bibliographystyle{ieee_fullname}
\bibliography{ref}


\renewcommand\thesection{\Alph{section}}
\renewcommand\thefigure{\Alph{figure}}  
\renewcommand\thetable{\Alph{table}}
\setcounter{section}{0}
\setcounter{figure}{0}

\clearpage
\begin{center}
    \section*{Supplementary Material}
\end{center}

In this material, we add more visualization results from the Places2 dataset\cite{zhou2017places} and CelebA dataset\cite{liu2015deep}. Besides, to validate the effectiveness of semantic filtering, we also visualize three specific channels of  $\mathbf{F}_3$ and $\hat{\mathbf{F}}_3$. Based on the feature level visualization results, we can see that semantic filtering can fill the large missing area with meaningful semantic information. All of these visualization results demonstrate the advantage of our method over the baseline methods.

\section{More Visualization Results}

\begin{figure*}
\centering
\includegraphics[width=1.0\linewidth]{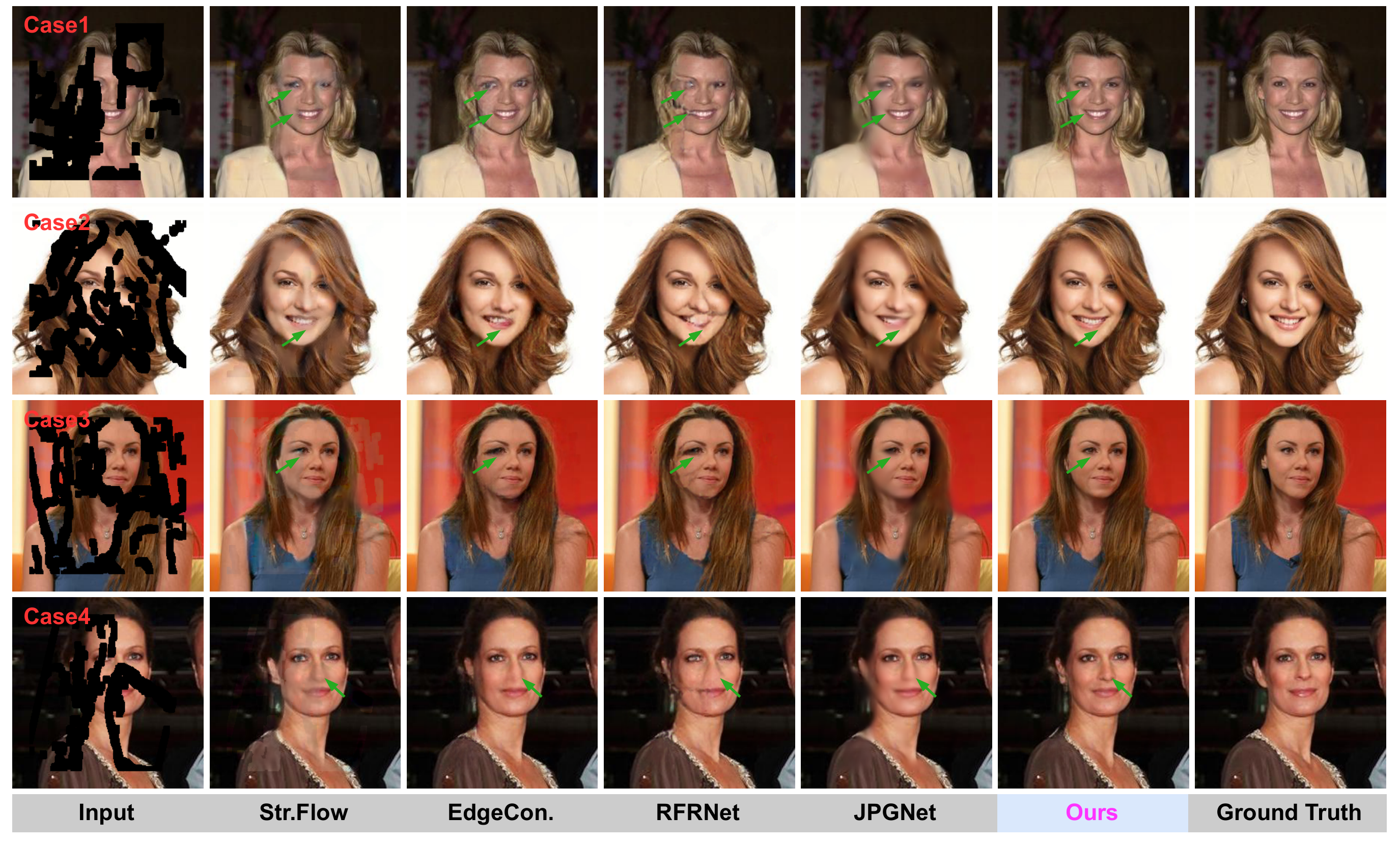}
\vspace{-10pt}
\caption{ Four visualization results of Str.Flow\cite{ren2019structureflow}, EdgeCon.\cite{nazeri2019edgeconnect}, RFRNet\cite{li2020recurrenta}, JPGNet\cite{guo2021jpgnet}, and ours method come from the CelebA dataset. We highlight the main differences via green arrows.
}
\label{fig:celebA_img}
\vspace{-5pt}
\end{figure*}

\begin{figure*}[t]
\centering
\includegraphics[width=1.0\linewidth]{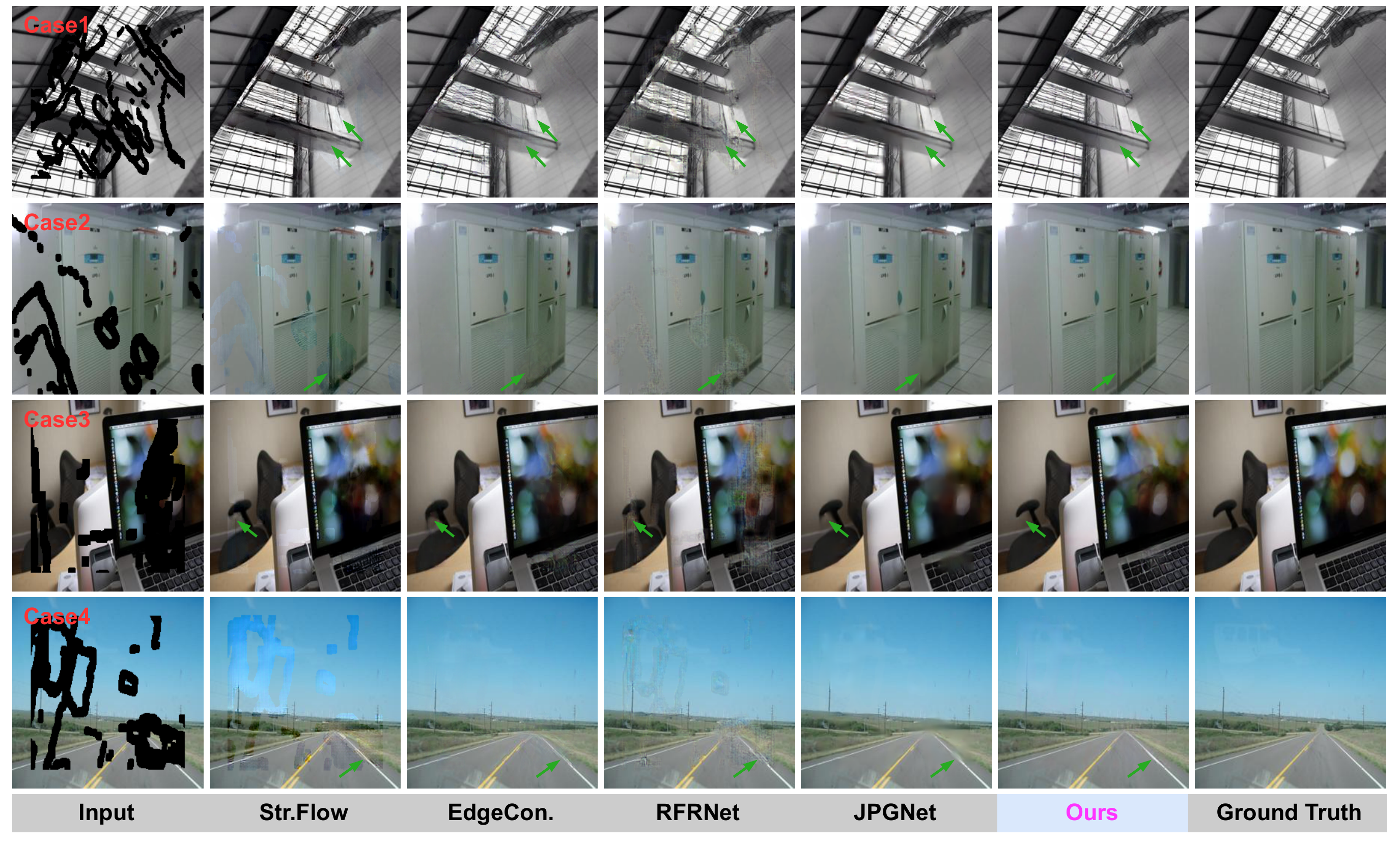}
\vspace{-10pt}
\caption{ Four visualization results of Str.Flow\cite{ren2019structureflow}, EdgeCon.\cite{nazeri2019edgeconnect}, RFRNet\cite{li2020recurrenta}, JPGNet\cite{guo2021jpgnet}, and ours method come from the Places2 dataset. We highlight the main differences via green arrows.
}
\label{fig:places2_img}
\vspace{-7pt}
\end{figure*}

We display more visualization results of Places2 dataset\cite{zhou2017places} and CelebA dataset\cite{liu2015deep} in \figref{fig:places2_img} and \figref{fig:celebA_img}, respectively.
As done in the submitted paper, we highlight the main differences across all inpainting results via green arrows. 
As the four cases shown in \figref{fig:celebA_img}, we can see that our method achieves the most natural and highest fidelity inpainting results with different faces and masks, which demonstrates the high generalization capability of our method. 
For example, in \figref{fig:celebA_img} case2, even the missing area are significantly large, the generated face is significantly close to the ground truth. In contrast, other methods introduce many artifacts like structure distortions and blur.

Moreover, for natural images in Place2 dataset \cite{zhou2017places}, our method is also able to recovery the detailed structures across different scenes that are very different from each other. Nevertheless, the baseline methods usually lead to distorted structures and unreal artifacts. For example, for the case1 in \figref{fig:places2_img}, Str.Flow, EdgeCon. \cite{nazeri2019edgeconnect}, and RFRNet distort the window and wall while JPGNet introduces slight blur. In contrast, our method can generate accurate window and wall structures that are very close to the ground truth. We have similar observations in other three cases.

%
%

\section{Visualization of Semantic Filtering}
To validate our semantic filtering at deep feature level, we visualize three specific channels of $\mathbf{F}_3$ and $\hat{\mathbf{F}}_3$ and compare them with the features of the naive encoder-decoder-based method. 

\begin{figure*}
\includegraphics[width=0.97\linewidth]{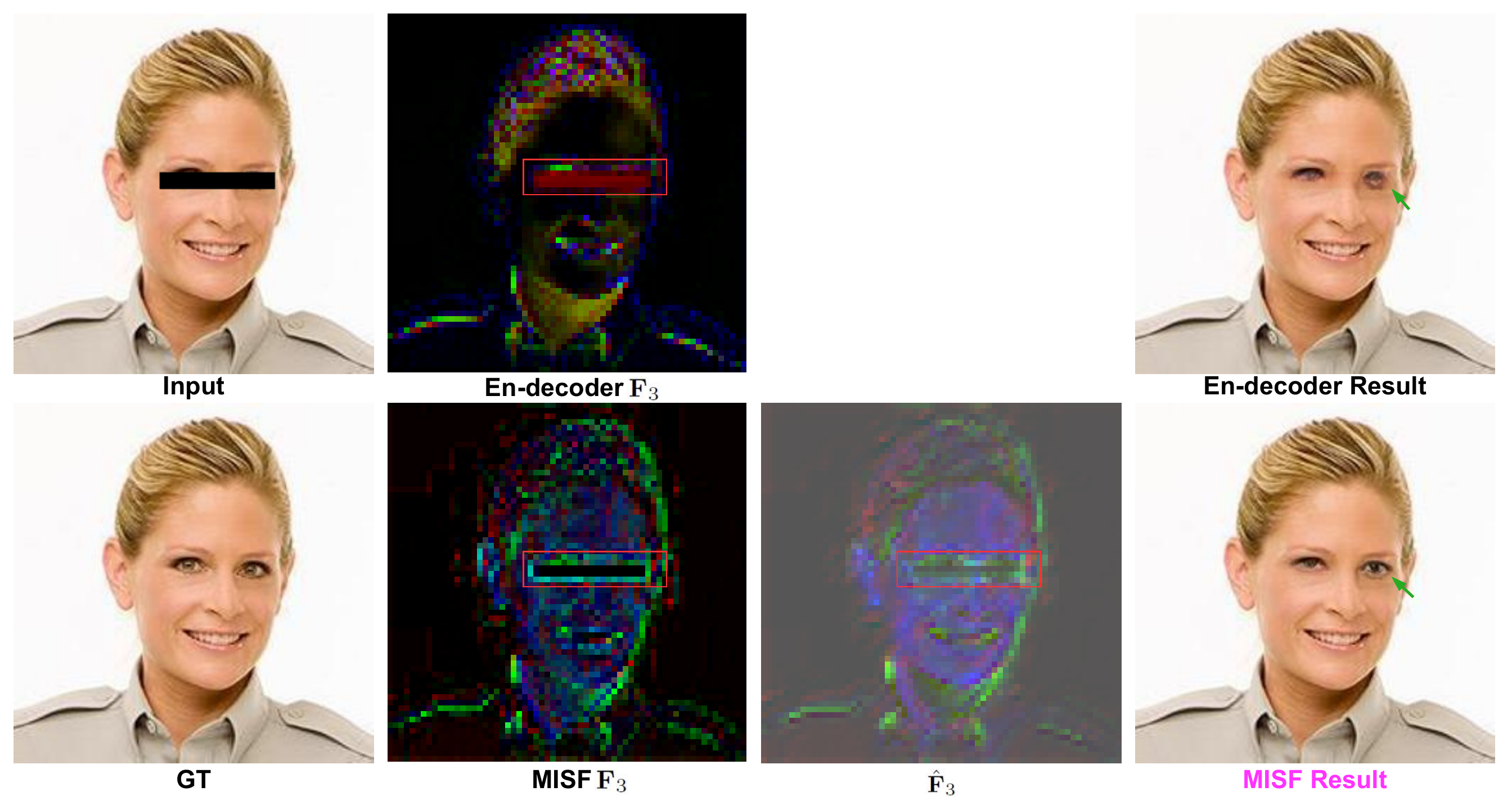}
\vspace{-5pt}
\caption{ Feature visualization result of case from the CelebA dataset. We highlight the feature level difference with red box and image level difference with green arrows.
}
\label{fig:f_1}
\vspace{-10pt}
\end{figure*}

\begin{figure*}
\includegraphics[width=0.97\linewidth]{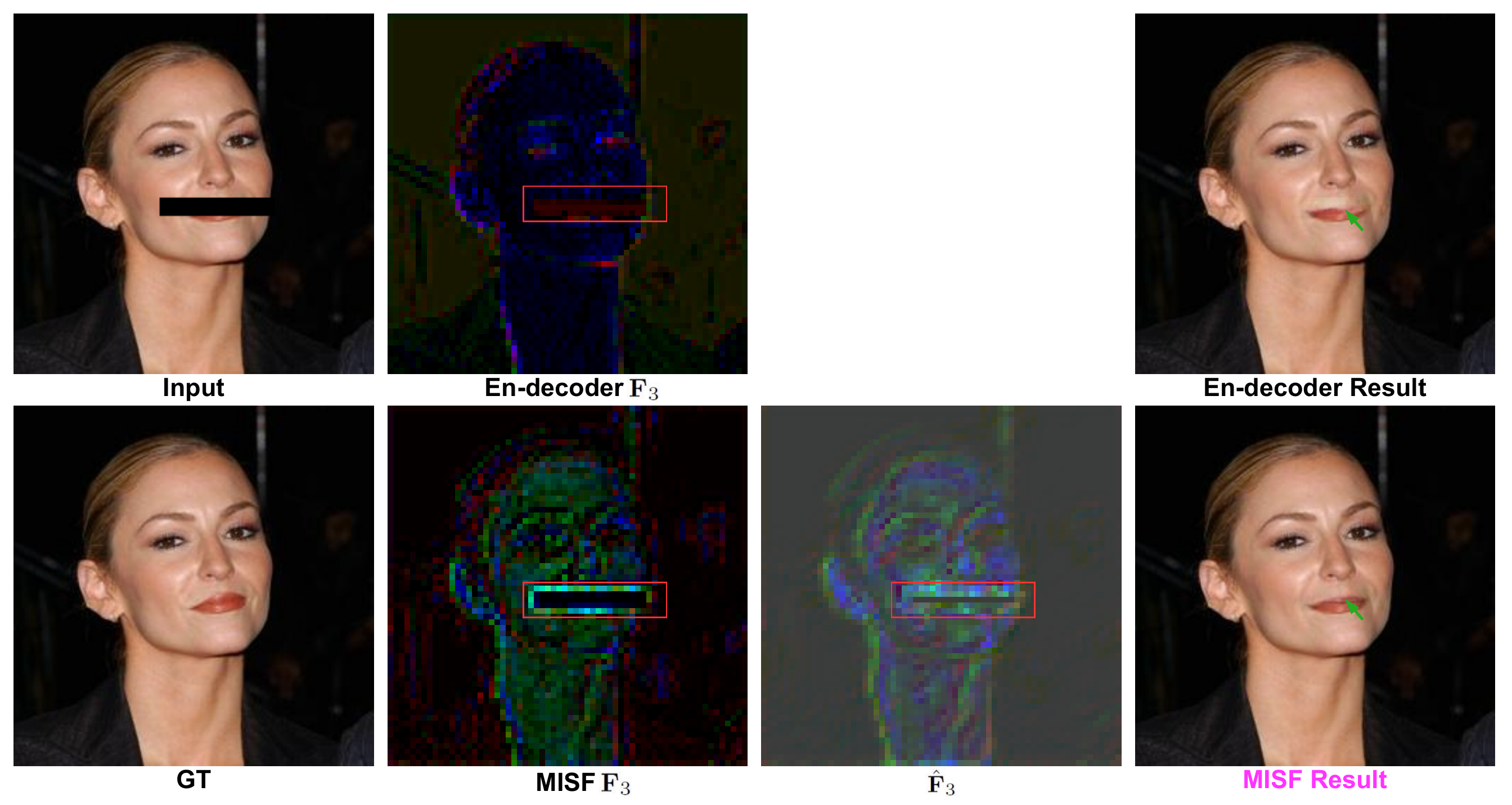}
\vspace{-5pt}
\caption{ Feature visualization result of case from the CelebA dataset. We highlight the feature level difference with red box and image level difference with green arrows.
}
\label{fig:f_2}
\vspace{-10pt}
\end{figure*}

\begin{figure*}
\centering
\includegraphics[width=0.97\linewidth]{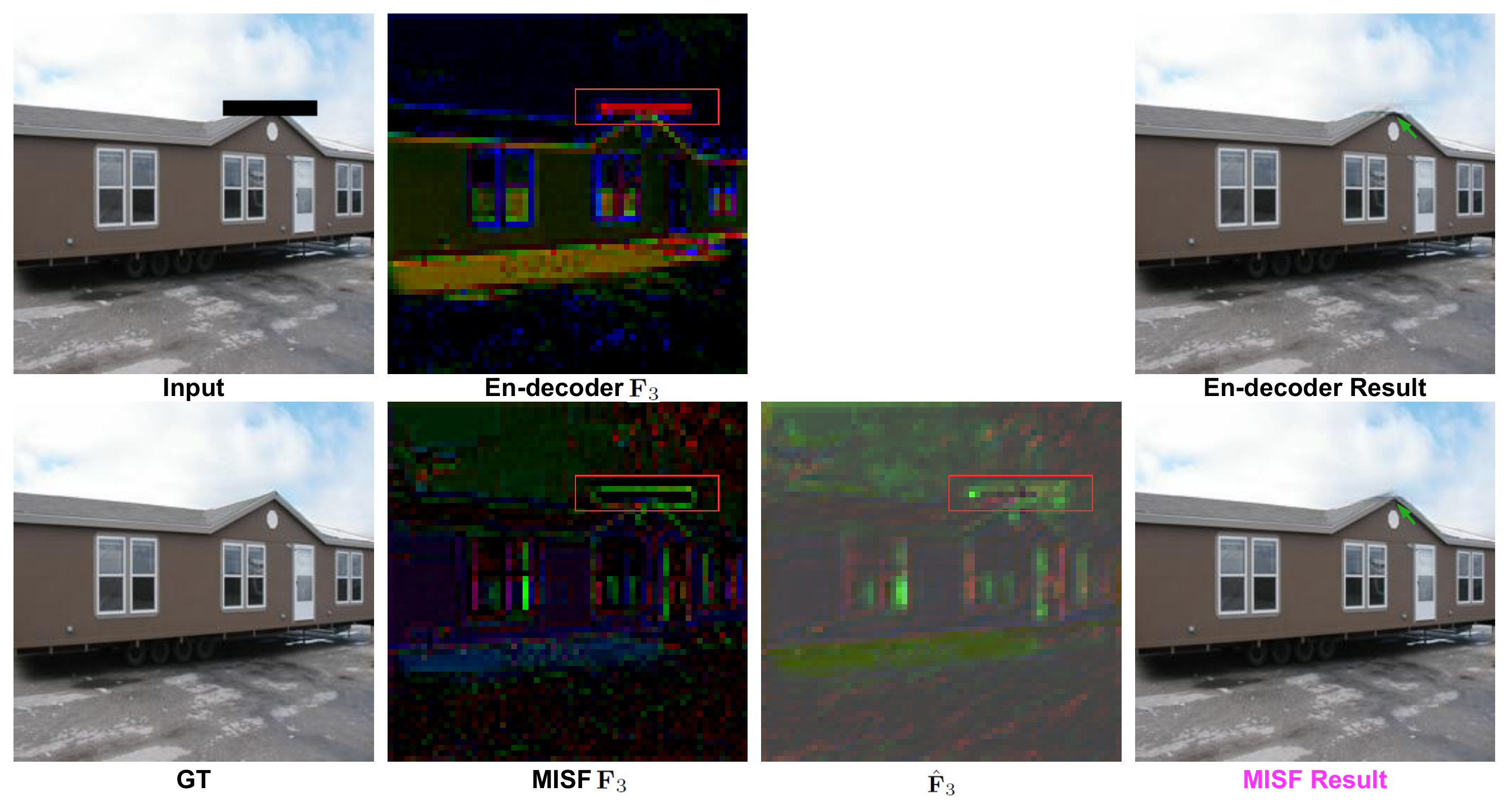}
\vspace{-5pt}
\caption{ Feature visualization result of case from the Places2 dataset. We highlight the feature level difference with red box and image level difference with green arrows.
}
\label{fig:f_3}
\vspace{-10pt}
\end{figure*}

\begin{figure*}
\includegraphics[width=0.97\linewidth]{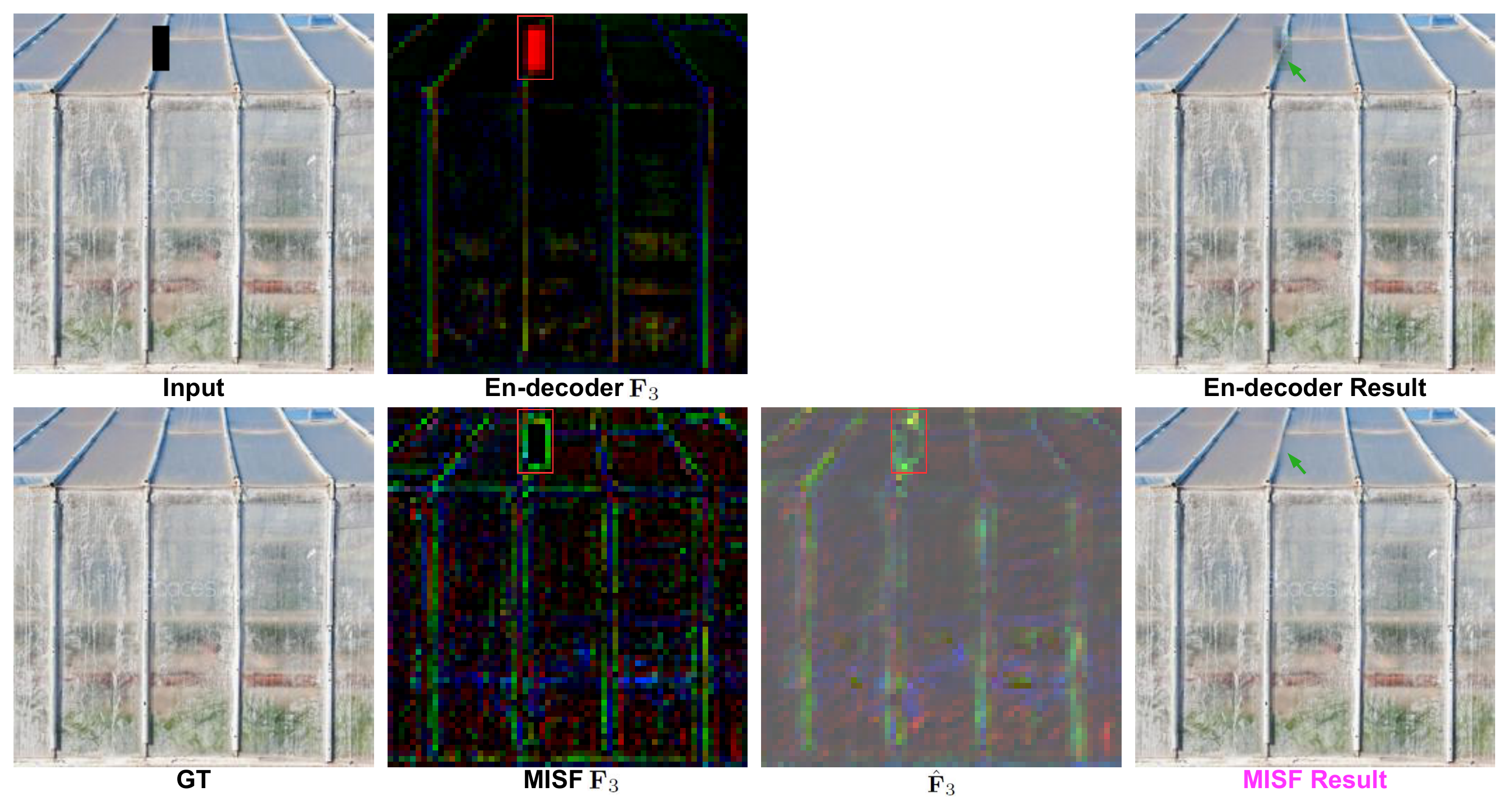}
\vspace{-5pt}
\caption{ Feature visualization result of case from the Places2 dataset. We highlight the feature level difference with red box and image level difference with green arrows.
}
\label{fig:f_4}
\vspace{-10pt}
\end{figure*}

As shown in \figref{fig:f_1} and \figref{fig:f_2}, when we crop the eye and mouth areas, the deep features of encoder-decoder-based method are significantly corrupted. As a result, the recovered face has distorted eyes. In contrast, our method is able to fill the miss semantic information at feature level (Compare the red box areas of $\mathbf{F}_3$ and $\hat{\mathbf{F}}_3$ in \figref{fig:f_1} and \figref{fig:f_2}). The final result of our method presents very clear and natural eyes with more details.
For natural scenes, we have similar observations in \figref{fig:f_3} and \figref{fig:f_4}. Specifically, in \figref{fig:f_3}, our method is able to fill more missing areas. in \figref{fig:f_4}, the encoder-decoder-based method is not able to recovery the structure and gets a distorted line. In contrast, our method can produce high-fidelity structures.

\end{document}